\documentclass{article}

\usepackage[final]{nips_2017}

\usepackage[utf8]{inputenc} 
\usepackage[T1]{fontenc}    
\usepackage{hyperref}       
\usepackage{url}            
\usepackage{booktabs}       
\usepackage{amsfonts}       
\usepackage{nicefrac}       
\usepackage{microtype}      
\usepackage{graphicx}

\let\svthefootnote\thefootnote
\newcommand\freefootnote[1]{%
  \let\thefootnote\relax%
  \footnotetext{#1}%
  \let\thefootnote\svthefootnote%
}

\title{Detecting hip fractures with radiologist-level performance using deep neural networks}

\author{
  William Gale\footnotemark, Gustavo Carneiro \\
  School of Computer Science \\
  The University of Adelaide\\
  Adelaide, SA 5000 \\
  \texttt{will@wgale.com} \\
  \texttt{gustavo.carneiro@adelaide.edu.au}\\
  \And
  Luke Oakden-Rayner\footnotemark[\value{footnote}], Lyle J. Palmer \\
  School of Public Health\\
  The University of Adelaide\\
  Adelaide, SA 5000 \\
  \texttt{\{luke.oakden-rayner,lyle.palmer\}} \\
  \texttt{@adelaide.edu.au}\\
  \And
  Andrew P. Bradley \\
  Faculty of Science and Engineering \\
  Queensland University of Technology \\
  Brisbane, QLD 4001 \\
  \texttt{a6.bradley@qut.edu.au} \\
}

\begin{document}
\freefootnote{* These authors contributed equally to the work}

\maketitle

\begin{abstract}
  We developed an automated deep learning system to detect hip fractures from frontal pelvic x-rays, an important and common radiological task. Our system was trained on a decade of clinical x-rays ($\approx$53,000 studies) and can be applied to clinical data, automatically excluding inappropriate and technically unsatisfactory studies. We demonstrate diagnostic performance equivalent to a human radiologist and an area under the ROC curve of 0.994. Translated to clinical practice, such a system has the potential to increase the efficiency of diagnosis, reduce the need for expensive additional testing, expand access to “expert level” medical image interpretation, and improve overall patient outcomes.
\end{abstract}

\section{Introduction}

Hip fractures represent a significant clinical and public health problem worldwide. They are among the most common causes of hospitalisation, morbidity, and mortality\textsuperscript{1} in the elderly, with a lifetime risk of 17.5\% for women and 6\% for men\textsuperscript{2}. The all-cause mortality rate is over 20\% within one year, and less than 50\% of patients regain the ability to live independently\textsuperscript{3}. 

Diagnosis of a fracture is usually made with pelvic x-ray imaging, and such imaging accounts for 6\% of all imaging referrals from the emergency department at our institution, a tertiary public hospital. To limit misdiagnosis, 5-10\% of at-risk patients undergo further imaging, including additional x-rays, nuclear medicine bone scans, computed tomography (CT), or magnetic resonance imaging (MRI), of which only a third demonstrate a fracture\textsuperscript{4}. Not only does this increase diagnostic costs and resource utilisation, but without access to these advanced imaging modalities (for example in remote and under-serviced regions) delayed or missed diagnosis is likely to result in worse patient outcomes including increased mortality rate\textsuperscript{5}, length of hospitalisation\textsuperscript{6}, and cost of care\textsuperscript{7}.
 
Recent advances in medical image analysis using deep learning\textsuperscript{8} have produced automated systems that can perform as well as human experts in some medical tasks\textsuperscript{9,10}. Deep learning is a computer science method that can be used to teach computers to recognise patterns that are useful in discriminating between groups of images, such as images with or without a certain disease\textsuperscript{8}. 

Highly sensitive and specific automation of hip fracture assessment using x-ray studies would lead to earlier and more accurate diagnosis and hence improve patient outcomes. Such automation would also reduce the need for expensive CT and MRI studies, which could improve service efficiency and increase access to highly accurate detection of hip fractures in under-serviced regions. Automation could also improve reproducibility, given the reported variation in diagnostic certainty among human experts of different experience levels\textsuperscript{11}. Here we investigate the application of deep learning using convolutional neural networks (CNNs) for the task of fracture detection, and present the first large scale study where a deep learning system achieves human-level performance on a common and important radiological task.

\section{Dataset}
\subsection{Developing ground truth labels for hip fractures}
Hip fractures are a promising target for machine learning approaches because of the availability of near-perfect ground truth labels. Clinically, patients with hip fractures do not remain undetected. Because of the weight bearing nature of the region, clinically 'silent' fractures rapidly progress to severe pain and immobility. As such, all patients with hip fractures that have imaging in a hospital should be identified in the radiology reports, the orthopaedic operative records, or the mortality records. While a small subset of patients may be lost to the records (due to hospital transfers, for example), our clinical experience suggests that the ground-truth label accuracy will be \textgreater 99\%.

\subsection{Obtaining and efficiently labelling our dataset}
Data for this study were obtained from the clinical radiology archive at the The Royal Adelaide Hospital (RAH), a large tertiary teaching hospital. Ethics approval was granted by the RAH human research ethics committee. All pelvis x-rays between 2005 and 2015 were included in the study, obtained with a wide variety of equipment used across the decade during normal clinical practice. Initial fracture labels were obtained by combining the orthopaedic surgical unit records, and findings from the radiology report archive (using regular expressions). This combination of sources resulted in labelling accuracy of around 95\%, evaluated on the hold out test set (described below) which was labelled manually by a radiologist using all of the available sources of information. To improve this further while avoiding the need for manual review of the entire dataset, a deep learning model was trained on the original labels and the false positive cases that this model identified were reviewed by a radiologist. As the "default" label for a case was negative (i.e., the case was not present in the orthopaedic database, nor had matching keywords in the radiology report), the majority of label errors were unrecognised fractures (which a well-trained model identifies as false positives). This process was repeated several times, and finally a single review of the false negatives was performed. This process improved the label accuracy on the hold-out test set from around 95\% to \textgreater 99.9\%, while only requiring 3371 cases to be labeled by an expert (7.4\% of the dataset). All manual review of cases was performed by a consultant radiologist (Dr Oakden-Rayner).

Each case provided two images, one from each hip, resulting in a total dataset of 53,278 images. These were randomly divided into a training set (45,492 images), a validation set (4,432 images) for model selection, and a held-out test set (3,354 images). There was no overlap of patients between the sets. The test set included only images referred from the emergency department (ED), which was considered the most clinically challenging setting, where lateral films and cross-sectional imaging are often not immediately available and management is often required prior to a formal radiology report. The prevalence of fractures among the patients in the test set was 19\%, which is the clinical prevalence in the ED at our centre. The prevalence of fractures among patients in the training and validation sets was lower ($\approx$12\%) due to the presence of outpatient and inpatient cases.

To ensure an unbiased set of labels for model evaluation, the entire test set was reviewed manually. 

\subsection{Managing medical data heterogeneity}
To deal with the variation inherent in clinical studies, the dataset was processed with a series of artificial neural networks. Firstly, a small CNN (CNN-frontal) was trained to identify frontal pelvis x-rays within each “case”, which often include other images like lateral hip films, chest, and spinal x-rays, requiring the network to learn to discriminate between gross anatomical features. Secondly, a regression-based CNN (CNN-bounding) was trained to localise the neck of femur, which is the only location where a relevant fracture could occur. This reduced the input size of the x-rays from over 3000 x 3000 pixels to a much more manageable 1024 x 1024 pixels, while maintaining image resolution. This task is more challenging than that of CNN-frontal, as it requires the system to localise fine-grained anatomical landmarks. Finally, a third CNN (CNN-metal) was trained to exclude cases with implanted metal from hip fractures and other similar operations (which represent a separate diagnostic challenge). Each network was trained on a small volume of data, requiring less than one hour of annotation effort by a radiologist. In Table 1 we present the performance of these models on unseen validation data. The accuracy of CNN-bounding is estimated by manual review of a held out test set, where adequacy was defined as coverage including the femoral head, and the greater and lesser trochanters.

\begin{table}[t]
  \begin{center}
  \begin{tabular}{lllllll}
    \toprule
    \cmidrule{1-7}
    Model  & Training set &Validation set  & Precison & Recall & Accuracy & Parameters \\
    \midrule
    CNN-frontal & 581 cases & 300 cases & \textgreater 0.99 & \textgreater 0.99  & \textgreater 0.99 & 2,408\\
    CNN-bounding & 300 cases & 440 cases & - & - & 0.97 & 746,624\\
    CNN-metal  & 5330 cases   & 300 cases & \textgreater 0.99 & 1.0 & \textgreater 0.99 & 11,792 \\
    \bottomrule
  \end{tabular}
  \end{center}

Table 1:  Performance of pre-processing deep learning models. CNN-frontal: identifies frontal pelvic x-rays, and excludes all other images. CNN-bounding: localises the neck of femur region for extraction. CNN-metal: excludes cases with metal in the region of interest. The CNN-metal training set was identified using regular expressions to find appropriate keywords from the radiology reports. 
\end{table}

\section{Methods}
\subsection{Model selection}
To analyse the pelvic x-rays, we applied a type of CNN known as a DenseNet\textsuperscript{12}. This architecture utilises extensive feed-forward connections between the layers, which is thought to improve feature propagation in these networks. 

The validation set was used to determine the following hyper-parameters (using a grid search strategy): the layer width (number of units per layer), the choice of activation function and leak rate, the use of a secondary loss function, the types and extent of data augmentation, the level of regularisation, and the learning rate.

The final network was 172 layers deep, with 12 features/units per layer (a total of 1,434,176 parameters). We used leaky relu\textsuperscript{13} non-linear activations with a leak rate of 0.5 and pre-activation batch normalisation\textsuperscript{14}. The model optimised two loss functions; a primary loss related to the presence or absence of fractures, and a secondary loss to learn more specific fracture location information (intra-capsular, extra-capsular, and no fracture). This was motivated by similar approaches which improved performance in previous work\textsuperscript{9,10}. 

We applied extensive data augmentation, consisting of small translations, rotations, and shears, as well as histogram matching to account for the residual variation in pixel values even after processing (a common issue with medical images). Each augmentation technique resulted in an absolute improvment of around 0.01 AUC on the validation set.

The network was regularised with a dropout\textsuperscript{15} rate of 0.2 and a weight decay rate of 1e-5. The network was trained via stochastic gradient descent using the Adam optimiser\textsuperscript{16}, with a learning rate of 0.0001. The network was trained for 25 epochs, with a batch size of 14. The system was trained using PyTorch\textsuperscript{17} on a on a workstation comprised of a hexa-core Intel i7-6850k processor, 64GB of DDR4 RAM and two 12GB NVidia Titan X Pascal graphics cards, resulting in a wall-clock training time of around 22 hours for the final model.

We also compared the performance of our model against a ten layer fully-convolutional CNN with 4,722,944 parameters, and pre-trained deep neural networks that required downsampling of the input images. The DenseNet had significantly higher performance on the validation data (absolute AUC improvement = 0.035 in both cases).

\subsection{Evaluation}
The algorithm was evaluated using a hold-out test set containing 3,354 images, with 348 fractures. We compared the performance of the algorithm against recently published work on automated hip fracture detection\textsuperscript{18}, as well as against the original radiology reports to achieve an estimate of human expert performance. 

\subsection{Estimating human performance}
Our task of identifying hip fractures using only a frontal pelvic x-ray is a common clinical one, as often the frontal film is the only test that is available to make this diagnosis. However, assessing human performance from radiology reports is confounded by the fact that other sources of information can be available at the time of reporting, and the role that this information plays in the diagnosis is usually unstated in the radiology report. Sources of such information could include discussions with clinicians (including physical examination and surgical findings), or other imaging such as lateral x-ray images or follow-up films (such as repeat x-ray, CT or MRI). At our centre the x-ray reports are often only finalised several hours after imaging, which allows radiologists to “peek” at any follow up imaging.

To provide the strongest possible baseline we generate an estimate of the “upper bound” of human performance. That is, we assume that \textit{no} unstated information was used to report the films. A report is only considered a human error if it provides the wrong diagnosis (later proven on follow-up), if it clearly states that additional information was required to makle the diagnosis (i.e., a lateral film), or it is unequivocal that further imaging will be necessary to make a diagnosis. When reports had vague wording or it was unclear if the radiologist was recommending further imaging, we treated the stated diagnosis as the result of the frontal x-ray only. For example, "there is a subtle fracture, consider CT to confirm" would be treated as a finding of a fracture, but "equivocal appearance, CT recommended" or "the fracture was only demonstrated on the lateral film" would both be considered a failure to make the diagnosis with the frontal film alone. 

\section{Results}
Figure 1 shows the receiver operator characteristic curve for our model on the hold-out test data. The human upper bound performance is also presented as a point in ROC space.
 
\begin{figure}[h]
  \centering
      \includegraphics[totalheight=8cm]{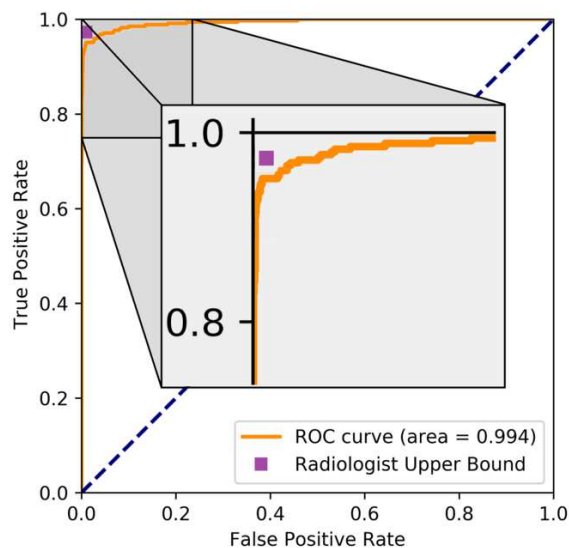}
  \caption{ROC curve showing the performance of the model with AUC 0.994, with a point reflecting the optimistic upper bound of human performance.}
\end{figure}

In Table 2 we compare our model to recently published results on automated hip fracture detection\textsuperscript{18}. Kazi et al. developed several models using a dataset of 669 frontal pelvic x-rays. They split these cases into two separate hip images, resulting in 900 images for training and 270 for testing. Their test data had a prevalence of 50\% and they presented results at a single operating point for each model. For consistency, we present results here with a similar data distribution (prevalence = 50\%) using all 348 fractures from our test set and 348 randomly selected non-fracture test cases. This subset was selected so the performance metrics were comparable (given the use of precision in Kazi et al., which varies with prevalence).

\begin{table}[t]
  \begin{center}
  \begin{tabular}{lllll}
    \toprule
    \cmidrule{1-5}
    Model  & Accuracy  & Precision & Recall & F1  \\
    \midrule
    Our model    & \textbf{0.97} & \textbf{0.99} & \textbf{0.95} & \textbf{0.97}      \\
    Kazi et al. (STN) & 0.84 (-13)  & 0.74 (-27) & 0.93 (-2) & 0.82 (-15)     \\
    Kazi et al. (LBM) & 0.81 (-16) & 0.76 (-25) & 0.84 (-11) & 0.80 (-17)    \\
    Kazi et al. (UBM) & 0.88 (-9) & 0.91 (-8) & 0.85 (-10) & 0.88 (-9)    \\
    \bottomrule
  \end{tabular}
  \end{center}
Table 2: Comparison against recent research. STN: spatial transformer network. LBM: lower bound model. UBM: upper bound model. The upper bound model presented by Kazi et al. required human localisation of the neck of femur region, our system performs this localisation automatically.
\end{table}

In Table 3 we present results for the entire larger test set (348 fractures and 2997 non-fractures), comparing our results against the performance on the original radiology reports. Note that in these tests the radiologists often had access to more information than our models, including any lateral x-rays, clinical information and follow-up studies. We present the performance of our system at two operating points, with high precision and high recall. Confidence intervals are calculated using exact Clopper-Pearson intervals\textsuperscript{20}.

\begin{table}[t]
  \begin{center}
  \begin{tabular}{lllll}
    \toprule
    \cmidrule{1-5}
    Model  & Acc (CI\textsubscript{95\%})  & Prec (CI\textsubscript{95\%})  & Rec (CI\textsubscript{95\%})  & F1 (CI\textsubscript{95\%})   \\
    \midrule
	Radiologist (estimate) &	\textbf{0.99} (99-100) &	0.93 (90-95) &	\textbf{0.97} (95-99) &	\textbf{0.95} (93-97) \\
	Our results (high prec) &	\textbf{0.99} (99-100) &	\textbf{0.97} (95-99) &	0.92 (89-95) &	\textbf{0.95} (93-97) \\
    Our results (high sens) &	\textbf{0.99} (99-100) & 0.92 (89-94) &	0.95 (92-97) &	0.94 (92-97) \\
    \bottomrule
  \end{tabular}
  \end{center}
Table 3: Comparison against estimated human baseline. The upper bound of human performance is estimated from the original (clinical) radiology reports for the test set cases. This reflects the upper limit of human performance given the data as we assume the radiologists used no additional information other than the frontal pelvic x-rays for their diagnoses (which is unlikely). As demonstrated, there is no significant difference between the human upper bound estimate and our model at either operating point on this dataset.
\end{table}

\section{Discussion}
We present a deep learning model that achieves human-level performance at the important radiology task of hip fracture detection in a large-scale study, and demonstrate state of the art performance compared to recently published work in this area by a large margin. The ROC AUC value of 0.994 is, to the best of our knowledge, the highest level ever reported for automated diagnosis in \textit{any} large-scale medical task, not just in radiology. Furthermore, unlike in previous work, our fully automated pipeline can take in any frontal pelvis x-ray and exclude ineligible cases, localise the neck of femur, and identify the presence of proximal femoral fractures automatically. Our research shows that despite the challenges specific to radiological image data, the development of large, clean datasets is sufficient to achieve high-level automated performance with deep-learning systems.

Our method of using small CNNs trained on small labeled datasets (300 cases for each model, taking less than one hour for an expert to annotate each dataset) made processing clinical image data practical. Many of these 'simple' tasks such as anatomy localisation can be achieved with very high accuracy with only a modest expert time commitment. Likewise, our process of iterative labelling using partially trained neural networks (where an expert reviews the false positives and false negatives rather than the entire dataset) allowed us to improve label accuracy from around 95\% to over 99.9\% while only hand-labelling 7.4\% of the dataset.

We also show several further new results; that high performance (AUC = 0.994) can be achieved without pre-training of models (for example, on ImageNet data as demonstrated in Gulshan et al. and Esteva et al.\textsuperscript{9,10}), and that very deep networks are useful in this setting. While these results should be expected given the results of CNNs in image analysis more generally, we have anecdotally seen many discussions around these points and we hope our findings can help to answer these concerns.

Baseline human performance on this task is difficult to estimate due to the presence of unreported external sources of information which can contribute to diagnostic decisions, and the use of inexact language by radiologists. While we are confident that our upper bound estimate of human performance is optimistic, and our team is currently working to test our system against human experts directly in a controlled environment.

\section*{References}

\small

[1] Brauer, C. A., Coca-Perraillon, M., Cutler, D. M. \& Rosen, A. B. Incidence and mortality of hip fractures in the United States. Jama 302, 1573-1579 (2009).

[2]	Kannus, P. et al. Epidemiology of hip fractures. Bone 18, S57-S63 (1996).

[3]	Morrison, R. S., Chassin, M. R. \& Siu, A. L. The medical consultant's role in caring for patients with hip fracture. Annals of internal medicine 128, 1010-1020 (1998).

[4]	Cannon, J., Silvestri, S. \& Munro, M. Imaging choices in occult hip fracture. The Journal of emergency medicine 37, 144-152 (2009).

[5]	Shiga, T., Wajima, Z. i. \& Ohe, Y. Is operative delay associated with increased mortality of hip fracture patients? Systematic review, meta-analysis, and meta-regression. Canadian Journal of Anesthesia 55, 146 (2008).

[6]	Simunovic, N., Devereaux, P. \& Bhandari, M. Surgery for hip fractures: Does surgical delay affect outcomes? Indian journal of orthopaedics 45, 27 (2011).

[7]	Shabat, S. et al. Economic consequences of operative delay for hip fractures in a non-profit institution. Orthopedics 26, 1197-1199 (2003).

[8]	LeCun, Y., Bengio, Y. \& Hinton, G. Deep learning. Nature 521, 436-444 (2015).

[9]	Gulshan, V. et al. Development and validation of a deep learning algorithm for detection of diabetic retinopathy in retinal fundus photographs. JAMA 316, 2402-2410 (2016).

[10]	Esteva, A. et al. Dermatologist-level classification of skin cancer with deep neural networks. Nature 542, 115-118 (2017).

[11]	Collin, D., Dunker, D., Göthlin, J. H. \& Geijer, M. Observer variation for radiography, computed tomography, and magnetic resonance imaging of occult hip fractures. Acta Radiologica 52, 871-874 (2011).

[12]	Huang, G., Liu, Z., Weinberger, K. Q. \& van der Maaten, L. Densely connected convolutional networks. arXiv preprint arXiv:1608.06993 (2016).

[13]	Maas, A. L., Hannun, A. Y., and Ng, A. Y. (2013). Rectifier nonlinearities improve neural network acoustic models. In Proc. ICML, volume 30.

[14]	Ioffe, S. \& Szegedy, C. in International Conference on Machine Learning.  448-456.

[15]	Srivastava, N., Hinton, G. E., Krizhevsky, A., Sutskever, I. \& Salakhutdinov, R. Dropout: a simple way to prevent neural networks from overfitting. Journal of Machine Learning Research 15, 1929-1958 (2014).

[16]	Kingma, D. \& Ba, J. Adam: A method for stochastic optimization. arXiv preprint arXiv:1412.6980 (2014).

[17]	Paszke, A., Gross, S. \& Chintala, S.     (2017).

[18]	Kazi, A. et al. in International Workshop on Machine Learning in Medical Imaging.  70-78 (Springer).

\end{document}